\documentclass{article} % For LaTeX2e
\usepackage{iclr2023_conference,times}

\usepackage{hyperref}
\usepackage{url}
\usepackage{graphicx}
\usepackage{amsmath}
\usepackage{amsfonts}
\usepackage{booktabs}

\title{XAI for transparent wind turbine power curve models}

\author{%
  Simon Letzgus\\
  Machine Learning Group\\
  Technische Universität Berlin\\
  Berlin, 10587, Germany \\
  \texttt{simon.letzgus@tu-berlin.de} \\
}

\iclrfinalcopy

\begin{document}

\maketitle

\begin{abstract}
Accurate wind turbine power curve models, which translate ambient conditions into turbine power output, are crucial for wind energy to scale and fulfil its proposed role in the global energy transition. While machine learning (ML) methods have shown significant advantages over parametric, physics-informed approaches, they are often criticised for being opaque "black boxes", which hinders their application in practice. We apply Shapley values, a popular explainable artificial intelligence (XAI) method, and the latest findings from XAI for regression models, to uncover the strategies ML models have learned from operational wind turbine data. Our findings reveal that the trend towards ever larger model architectures, driven by a focus on test set performance, can result in physically implausible model strategies. Therefore, we call for a more prominent role of XAI methods in model selection. Moreover, we propose a practical approach to utilize explanations for root cause analysis in the context of wind turbine performance monitoring. This can help to reduce downtime and increase the utilization of turbines in the field.

% In general, we recommend reevaluating the test-set-performance-driven trend towards ever larger ML-model architectures for wind turbine power curve modelling and call for a more prominent role of XAI methods in model selection.
\end{abstract}

\section{Introduction}
The energy sector is responsible for the majority of global greenhouse gas emissions \citet{owidco2andothergreenhousegasemissions} and wind energy is to play a key role in its decarbonization \citet{council2022gwec}. Accurate wind turbine power curve models are important enablers for this transition. Coupled with meteorological forecasts, they are used for energy yield prediction \citet{Optis2019, Nielson2020} and thereby crucial for stable operation of electricity grids with high wind penetration. Moreover, they have been successfully utilized for wind turbine condition monitoring \citet{Kusiak2009, Schlechtingen2013_2, Buttler2013}, which reduces downtime and directly increases the amount of renewable energy in the electricity mix. 

Therefore, power curve modelling has received plenty of attention (see \citet{Sohoni2016} for a comprehensive review). While early approaches have mainly focused on parametric models based on physical considerations (e.g. \citet{Kusiak2009}), complex non-linear ML models have become the state-of-the-art today \citet{Methaprayoon2007, Schlechtingen2013_2,  Pelletier2016, Optis2019, Nielson2020}. However, the wind community has often uttered the need for more transparency and interpretability of ML approaches to be trusted and deliver actionable results \citet{Tautz-Weinert2016, Sohoni2016, Optis2019, Chatterjee2021, Barreto2021}.

As a response, researchers have started to utilize methods from the field of XAI \citet{DBLP:journals/pieee/SamekMLAM21, montavon2018methods,LapNCOMM19} (compare for example \citet{Chatterjee2021}). Several works have applied Shapley values, a popular XAI method thanks to its easy to use implementation \citet{DBLP:conf/nips/LundbergL17}, for power prediction \citet{tenfjord2020value, Pang22_shapleyPC} and turbine monitoring \citet{chatterjee2020vesselshap, mathew2022estimation, movsessian2022interpretable}. However, their out-of-the-box application of the method limits the insights to qualitative importance rankings of features. We address this issue and take the approach to a quantitative level by including the latest findings from research on XAI for regression models \citet{Letzgus22}. This allows to assess whether data-driven models learn physically reasonable strategies from operational wind turbine data and consequently draw conclusions for model selection. Furthermore, we highlight the benefits in turbine performance monitoring, where we decompose the deviation from an expected turbine output and assign it to the input features in a quantitatively faithful manner.

\section{Data and Methods}

\subsection{Data}
We use operational data from the Supervisory Control and Data Acquisition (SCADA) system of two 2 MW wind turbines and a meteorological met-mast, all located within the same site. The data set is openly accessible \footnote{https://opendata.edp.com} and covers a period of two years. Our pre-processing pipeline ensures the turbine is in operation ($P>0 kW$) and not affected by stoppages or curtailment (filters based on SCADA logs). Overall, this results in roughly 50.000 data points per turbine, which we temporally divide into train and test set (one full year each), as well as a validation set (20\% randomly sampled from training year).

\subsection{Models}

\textbf{IEC models:} we have implemented a physics-informed baseline model following the widely adopted international standard IEC 61400-12-1 \citet{IEC61400}. It consists of a binned power curve and the respective corrections for air density and turbulence intensity (more details in App. \ref{app:models_and_performance}).

\textbf{ML models:} we train three different ML-models that have successfully been applied to power curve modelling, each representing an established model class (for more details see App. \ref{app:models_and_performance}): 
\begin{enumerate}
    \item $\bf{Random Forest (RF)}$: well established data-driven baseline, used e.g. in \citet{Kusiak2009, janssens2016data}.
    \item $\bf{ANN_{small}}$: moderately sized, best performing ANN from \citet{Schlechtingen2013_2} with 2 layers (3,3 neurons) and sigmoid activation functions.
    \item $\bf{ANN_{large}}$: state-of-the-art, fully connected ReLU network from \citet{Optis2019} with 3 layers (100,100,50 neurons).
\end{enumerate}

As model inputs, we select wind speed ($\mathbf{v_w}$), air density ({\boldmath$\rho$}) and turbulence intensity ($\mathbf{TI}$). This limits complexity and enables a fair comparison with the physical baseline model (both, in terms of performance and strategy). RFs were optimized using CART, the ANNs using Adam with an initial, adaptive learning rate of 0.1 and early stopping regularization after waiting for 100 epochs. We used the respective implementations of the {\tt scikit-learn} library \citet{scikit-learn}. Table \ref{tab_model_performance} gives an overview of the implemented models and their performance for the two turbines. As expected, the ML models outperform the IEC model. In line with literature, the large ANNs show best performance followed by small ANNs and the RFs across all settings \citet{Sohoni2016}.

\begin{table}[h]
\caption{Summary of model performance ($RMSE_{test}$ [kW])}
\label{tab_model_performance}
\centering
    \begin{tabular}{l c  c} 
        \toprule
         Model & Turbine A & Turbine B\\
        \hline
        IEC model       &  43.60         & 35.40 \\
        % \hline
        $RF$              & $37.24\pm 0.02$  & $34.42\pm0.03$\\
        % \hline
        $ANN_{small}$     & $35.92\pm 0.65$ & $33.34\pm0.66$\\
        % \hline
        $ANN_{large}$     & $\textbf{35.36}\pm0.51$   & $\textbf{32.88}\pm0.37$ \\ 
        \bottomrule
    \end{tabular}
\end{table}

\subsection{Meaningful Shapley values for power curve models}
\textit{Shapley values} \citet{Shapley1953} determine the contribution of a feature by removing it and observing the effect, averaged over all permutations \citet{DBLP:conf/nips/LundbergL17}. Its conservation property, combined with appropriately chosen reference points ($\widetilde{x}$), enables quantitatively faithful attributions which retain the unit of the model output \citet{Letzgus22}. For wind turbine power curve models we advocate domain-specific settings of $\widetilde{x}$ rather than the commonly used $\widetilde{x}_{mean} = \bar{X}_{tr}$ \citet{mathew2022estimation, movsessian2022interpretable, tenfjord2020value, Pang22_shapleyPC} (see App. \ref{app:x_ref}). For assessing physical compliance of model strategies, we generate attributions for both, physical and data-driven models, and compute correlation coefficients between them ($\bf{R^2_{phys}}$). This has shown to be a suitable quantitative indicator for the extent to which data-driven models follow the expected fluid mechanical principles (Sec. \ref{sec:results_strategy}).

\section{Results}

\subsection{Do ML models learn physically reasonable strategies?}
\label{sec:results_strategy}

\begin{figure}[t]
    \centering
    \includegraphics[width=0.975\linewidth]{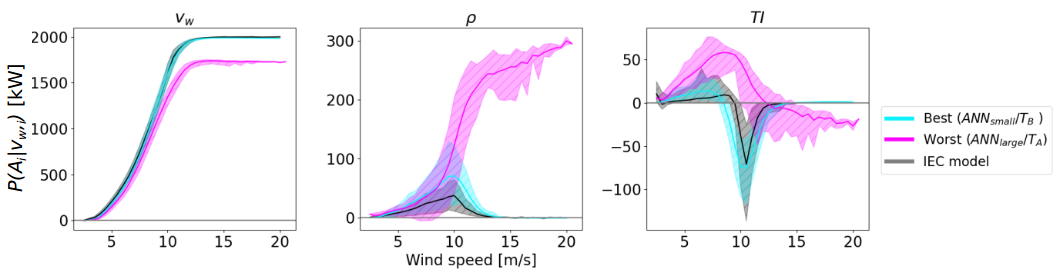}
    \caption{Distributions of attributions (mean as lines, range shaded) conditioned on wind speed ($v_w$) by feature (left, center, right). They represent the strategies that agree most (cyan) and least (pink) with the physical benchmark model (grey), and exemplify the wide range of of strategies learned.}
    \label{fig:strategies}
\end{figure}

\begin{figure}[b]
    \centering
    \includegraphics[width=0.975\linewidth]{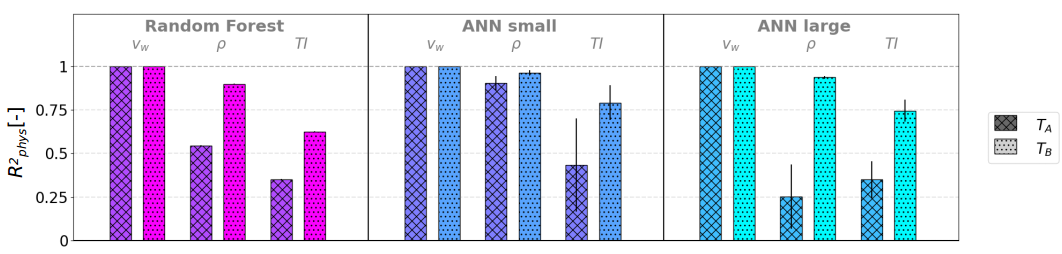}
    \caption{{Mean correlation between data-driven and physical IEC model attributions by model (sections/color schemes), input parameter (subsections) and turbine (shade/hash). Error bars to the standard deviation of correlation over 10 training runs with different initializations.}}
    \label{fig:res_boxnbars}
\end{figure}

As so often, the answer to this question is: it depends. Our results have revealed that ML models can learn a surprisingly wide range of strategies from operational SCADA data, some of which capture physical relationships in an almost textbook-like manner and others that fail to consider them almost entirely. Figure \ref{fig:strategies} facilitates a comparison between these extreme cases by visualizing strategies for the model with highest ($R^2=0.95$ - $ANN_{small}/T_{B}$) and the lowest ($R^2=0.47$ - $ANN_{large}/T_{A}$) correlation with the physics informed baseline (also displayed). To understand the reasons, we conduct a systematic analysis of potential impact factors. Figure \ref{fig:res_boxnbars} shows the similarity of learned strategies with the physical baseline for the different models, turbines and input features.

\textbf{Strategies by input features:} intuitive ranking of physical compliance that coincides with the physical importance of the respective input features (influence of $v_w$ captured best, $\rho$ mostly reasonable, and $TI$ worst).

\textbf{Impact of model:} $ANN_{large}$  exhibits lower average agreement with IEC model strategies than $RF$  or $ANN_{small}$. The latter learned the most physical strategies across all settings. Also, ANN model initialization can have a profound impact on the learned strategy (see 'error bars' in Fig. \ref{fig:res_boxnbars}).

\textbf{Impact of turbine:} All ML models show higher overall agreement with IEC model strategies on $T_B$, large models in particular. The IEC model itself also performs much better on $T_A$, indicating that the data set contains less effects beyond the considered physical phenomena.

\textbf{Implications for model selection:} Physically reasonable power curve models are desirable and more robust in out of distribution settings (see App. \ref{app:models_and_performance}). Unfortunately, test-set RMSE is not a good indicator for physical compliance. $ANN_{large}$, for instance, show best performance but largest deviation from physical strategies. Based on our results, we recommend the use of moderately sized ANNs instead (similar to $ANN_{small}$). They have shown clear advantages in terms of strategy at only slightly increased test-set errors compared to $ANN_{large}$ while outperforming $RFs$ in both categories.

\subsection{Explaining deviations from an expected turbine output}
\label{sec:results_underperformance}
We demonstrate the importance of appropriate reference points for quantitatively faithful attributions and highlight their potential in the context of performance monitoring. 

\begin{figure}[b]
\centering
\includegraphics[width=\linewidth]{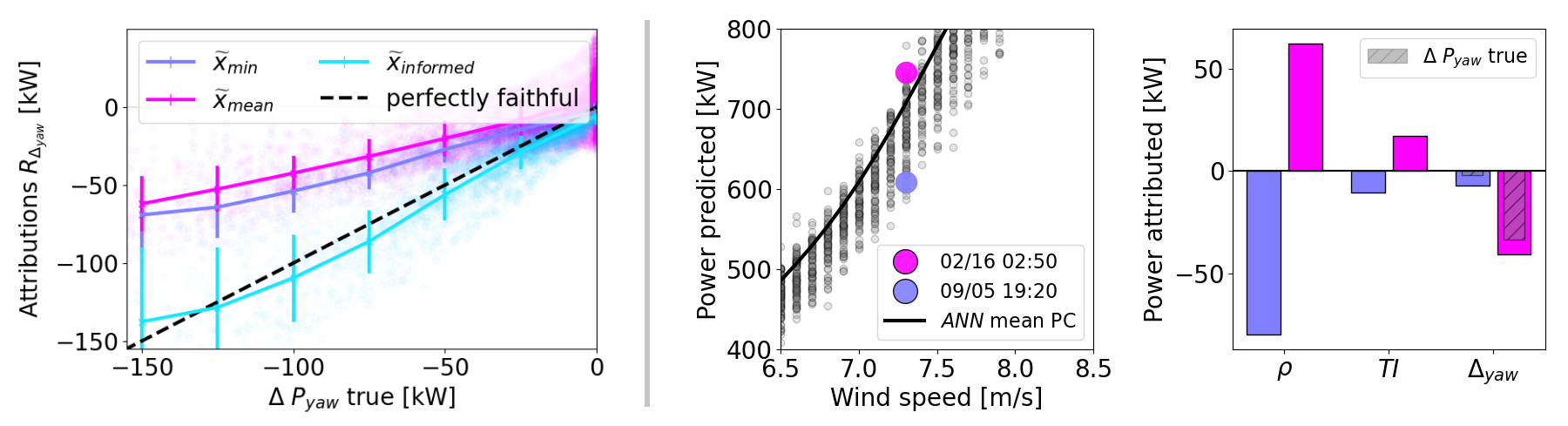}
\caption{\textbf{Left}: Faithfulness for different reference points ($\widetilde{x}$) - true yaw-induced output reduction vs. respective attributions. Lines display mean and standard deviations over 25 kW bins. \textbf{Right}: two selected data points relative to the learned standard power curve (ANN mean PC), and their respective attributions per input feature.}
\label{fig_expl_deviations}
\end{figure}

We utilize a model of type $ANN_{small}$ and include a yaw misalignment feature (details can be found in App. \ref{app:monitoring}). This enables experiments in a controlled fashion and a comparison between magnitudes of Shapley attributions ($R_{\Delta_{yaw}}$) and the ground truth  ($\Delta P_{yaw}$). We compare attributions for three different reference points (Fig. \ref{fig_expl_deviations}, left): $\widetilde{x}_{min}$ (used for global strategies earlier), $\widetilde{x}_{mean}$ (often the standard choice), and $\widetilde{x}_{informed}$, which incorporates the assumptions implicit to the expected output relative to which we explain (see App. \ref{app:x_ref}). The latter clearly outperforms both others in the presented experiment and is, therefore, recommended for this task.

Once we have ensured physically plausible model strategies and quantitatively faithful attributions, we can utilize the explanations in a performance monitoring context (Fig. \ref{fig_expl_deviations}, right). The first plot shows two selected data points and the learned power curve under mean ambient conditions (no yaw misalignment). Both instances have a low absolute error of $\sim$10 kW. Their position relative to each other and the mean curve suggests that the September instance (violet) is much more likely to be affected by yaw misalignment than the February instance (pink) (higher output at the same wind speed). The respective attributions, however, enable a decomposition and quantification of different entangled effects. In February, significant yaw misalignment was present but compensated by favourable ambient conditions. The September instance's lower output, on the other hand, can be attributed to less advantageous environmental conditions rather than a technical malfunction. While monitoring yaw misalignment directly would be a trivial solution in this example, this is usually not possible for potential other root causes (blade-pitch-angles or turbine interactions, for example) which can be identified with the approach. Nevertheless, the potential to indicate root causes for underperformance is naturally limited to effects that are correctly captured by the model in the first place. Additionally, the absolute model error should serve as a confidence measure regarding the corresponding attributions (explanations for data points with a low error are more trustworthy).

\section{Conclusions}

So far, the trend towards ever more complex ML models in data-driven power curve modelling was justified based on superior test-set performance. The results from this contribution suggest, that this might be on the cost of physically reasonable model strategies. These findings remind of the famous Clever-Hans effect in classification \citet{LapNCOMM19}. Whether the deviation from physical intuition can have similarly severe implications here, should be the subject of further research. 

Overall, it has to be considered whether a focus on smaller models, that better capture physical phenomena, would be the right way forward. Another benefit of smaller architectures is the reduced carbon footprint when training individual models for each turbine of a large fleet. In any case, we recommend a more prominent role of XAI methods in model selection, since often physically more plausible models were obtained with only minor performance losses. Moreover, we have introduced a practical approach to utilize XAI attributions in a quantitatively faithful manner. This is particularly useful in turbine performance monitoring where the increased transparency can help to identify root causes for underperformance.

With this work, we have laid the foundation for transparent data-driven wind turbine power curve models. We hope that the insights will help practitioners to more effectively utilise their ML models and turbines; along with the related positive implications for the global energy transition.

%\section*{References}
%\bibliographystyle{unsrt}

\bibliography{bibliography}
\bibliographystyle{iclr2023_conference}

\newpage

\appendix
\section{Domain and problem specific reference points}
\label{app:x_ref}
The current practice in the wind domain is to use the mean training input feature vector $\widetilde{x}_{mean}=\bar{X_{tr}}$ (compare e.g. \citet{mathew2022estimation, movsessian2022interpretable, tenfjord2020value, Pang22_shapleyPC}). We advocate domain-specific settings for the application to wind turbine power curve models: 

\textbf{Global explanations:} For explaining global model strategies, we suggest explaining relative to reference point $\widetilde{x}_{min}=min(X_{tr})$. This generates intuitive attributions relative to wind speed zero (or cut-in, depending on data pre-processing). Plotting their distributions conditioned on the measured wind speed $P(A_i | v_w)$ against the wind speed additionally resembles the way power curves are typically displayed (compare Fig. \ref{fig:strategies}). This facilitates contextualization and interpretation by domain experts. 

\textbf{Local explanations:} In the context of wind turbine power curves, the most common question is to explain a deviation from an expected turbine output. This requires $\widetilde{x}$ to reflect the implicit assumptions of that expectation. We suggest the informed reference point ($\widetilde{x}_{informed}$) to be conditioned on $v_w$: $x_{ref_i} = \mathbb{E}(x_i|v_w)$ for environmental parameters, and set to a healthy parameter baseline for technical parameters (e.g. zero for a yaw misalignment feature).

% \section{Data pre-processing}
% \label{app:preprocessing}
% Data pre-processing consists of two consecutive filters:
% \begin{enumerate}
%     \item Non-operational periods: turbine power output $< 0$ kW
%     \item Non-normal periods: removes data points affected by stoppages, ramp-ups and curtailment based on SCADA log-messages
% \end{enumerate}

\section{Models and performance}
\label{app:models_and_performance}

\textbf{IEC models:} For each turbine, we calculate the binned power curve and apply air density, as well as TI corrections following the IEC standard \citet{IEC61400}. All required parameterizations for the physics-informed model (binned power curve, average air density and zero TI-reference power curve) are calculated using the training data set. Performance evaluation and explanations are calculated for the test set. More details can be found in IEC 61400-12 \citet{IEC61400}.

\textbf{ML models:} hyperparameters which were not specified in the respective publications (training modalities, for example), were selected based on a gird-search with 5-fold cross-validation on the training data set. For evaluation we then report the test set RMSE (mean and standard deviation over 10 training runs with different model initializations). All parameters not further specified were left at the standard settings of the {\tt scikit-learn} \citet{scikit-learn} implementations: 

$\bf{RF:}$ \newline
{\tt RandomForestRegressor(min\_samples\_split=3, min\_samples\_leaf=30, n\_estimators=100)}

$\bf{ANN_{small}}$\citet{Schlechtingen2013_2}: \newline
{\tt MLPRegressor(hidden\_layer\_sizes=(3, 3), activation='logistic',
                             learning\_rate\_init=0.1, learning\_rate='adaptive', max\_iter=10000, tol=10**-6,
                             alpha=0, early\_stopping=True, n\_iter\_no\_change=100, verbose=0)}

$\bf{ANN_{large}}$ \citet{Optis2019}: \newline
{\tt MLPRegressor(hidden\_layer\_sizes=(100, 100, 25), activation='relu',
                             learning\_rate\_init=0.1, learning\_rate='adaptive', max\_iter=10000, tol=10**-6,
                             alpha=0, early\_stopping=True, n\_iter\_no\_change=100,
                             verbose=0)}

\begin{figure}[t]
    \centering
    \includegraphics[width=0.6\linewidth]{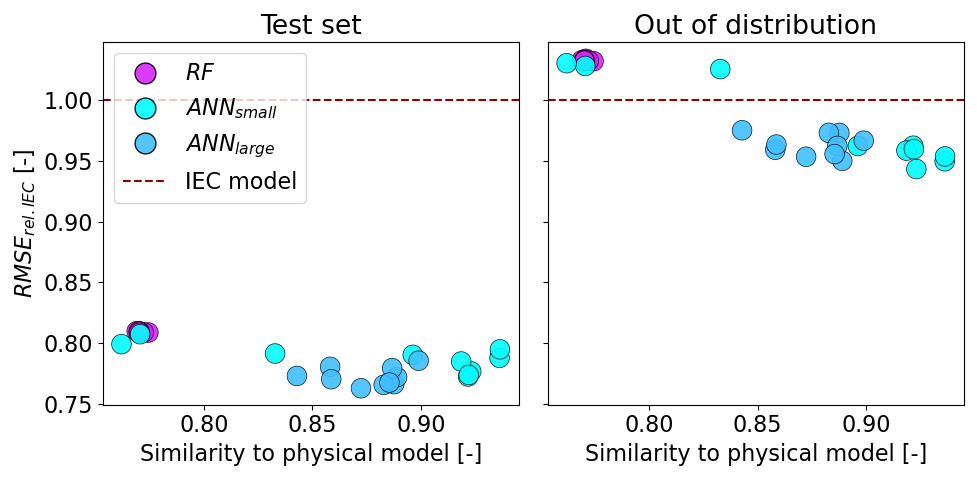}
    \caption{ML model strategy and vs. RMSE for test set (left) and an out of distribution test set (right). Each data point represents a model trained with a different initialization. Results are shown for $T_B$. }
    \label{fig:out_of_dist}
\end{figure}

Lastly, we have created an out of distribution scenario to analyse the relation of model strategy and performance. We apply an additional norm filter that removes data points that are further away than 100 MW from the manufacturers standard power curve. We then train the models on the filtered data and evaluate them on the filtered test data (Figure \ref{fig:out_of_dist}, left) and the data during the test period that was removed by the norm-filter (Figure \ref{fig:out_of_dist}, right). The results show, that physically plausible models are more robust in out of distribution scenarios. 

\newpage

\section{Data augmentation with yaw misalignment}
\label{app:monitoring}

We utilize a model of type $ANN_{small}$ and include the absolute difference between average wind- and nacelle direction as a yaw-misalignment feature ($\Delta_{yaw}$). This allows for experiments in a controlled fashion by augmenting data with artificial yaw-misalignment. This is achieved by adding normally distributed yaw misalignment of up to $\pm 15^\circ$ to our data sets, and multiplying the respective targets (turbine output) with a yaw-misalignment factor $c_{ymis,i} = cos^3(\Delta_{yaw})$, if $v_{w,i}<v_{rated}$ (approximation derived from the actuator disk model \citet{howland2020influence}). After training and evaluation of the model on the augmented data, we can compare magnitude of Shapley attributions to the ground truth:
\begin{equation}
    \Delta P_{yaw,i}true = 
    \begin{cases}
        c_{ymis,i} \cdot P_{T,i}, \,& \text{if } v_{w,i}<v_{rated}\\
        0,              \,& \text{otherwise}.
    \end{cases}
\end{equation}

\end{document}